\documentclass[11pt,a4paper]{article}
\usepackage[hyperref]{acl2019}
\usepackage{times}
\usepackage{latexsym}

\usepackage{url}

\aclfinalcopy 

\usepackage[textwidth=2.1cm]{todonotes} 

\newcommand{\jb}[1]{#1}
\newcommand{\la}[1]{#1}

\title{Thirty Musts for Meaning Banking}

\author{Johan Bos \\
  Center for Language and Cognition\\
  University of Groningen \\
  \texttt{johan.bos@rug.nl} \\\And
  Lasha Abzianidze \\
  Center for Language and Cognition\\
  University of Groningen \\
  \texttt{l.abzianidze@rug.nl} \\}

\date{}

\begin{document}
\maketitle

\begin{abstract}

Meaning banking---creating a semantically annotated corpus for the purpose of semantic parsing or generation---is a challenging task.  
It is quite simple to come up with a complex meaning representation, 
but it is hard to design a
simple meaning representation that captures many nuances of meaning.
This paper lists some lessons learned in nearly ten years of
meaning annotation during the development of the Groningen Meaning Bank \cite{gmb} and the Parallel Meaning Bank \cite{pmb}. 
The paper's format is rather unconventional: there is no
explicit related work, no methodology section, no results, and no
discussion (and the current snippet is not an abstract but actually an introductory
preface). Instead, its structure is inspired by work of \citet{Traum:20} and \citet{Bender:100}.
\la{The list starts with a brief overview of the existing meaning banks (Section\,\ref{sec:overview}) and the rest of the items are roughly divided into three groups: 
corpus collection (Section\,\ref{sec:select_corpora} and \ref{sec:freeze}, 
annotation methods (Section\,\ref{sec:raw}--\ref{sec:neutral_tools}),
and design of meaning representations (Section~\ref{sec:normalize_symbols}--\ref{sec:measure}).}
We hope this overview
will give inspiration and guidance in creating improved meaning banks in the future.

\end{abstract}

\section{Look at other meaning banks}\label{sec:overview}

Other semantic annotation projects can be inspiring, help you to find
solutions to hard annotation problems, or to find out where
improvements to the state of the art are still needed \cite{abend-rappoport-2017-state}.
Good starting points are the English Resource Grammar \cite{Flickinger:2000,Flickinger:2011},
the Groningen Meaning Bank (GMB, \citealt{gmb}), the AMR Bank \cite{amr},
the Parallel Meaning Bank (PMB, \citealt{pmb}), Scope Control Theory \cite{sct},
UCCA \cite{ucca}, \la{Prague Semantic Dependencies \cite{pragueDT:17}}
and the ULF Corpus based on Episodic Logic \cite{ulf}.
The largest differences between these approaches can be found in the
expressive power of the meaning representations used. The simplest
representations correspond to graphs \cite{amr,ucca}; slightly more
expressive ones correspond to first-order logic
\cite{Oep:Kuh:Miy:16,gmb,pmb,sct}, whereas others go beyond this
\cite{ulf}. Generally, an increase of expressive power causes a
decrease of efficient reasoning \cite{blackburnbos:2005}. Semantic
formalisms based on graphs are attractive because of their simplicity,
but will face issues when dealing with negation in inference tasks
(Section~\ref{sec:negation}).  The choice might depend on the
application (e.g., if you are not interested in detecting
contradictions, coping with negation is less important), but arguably,
an open-domain meaning bank ought to be independent of a specific
application.

\section{Select public domain corpora}\label{sec:select_corpora}

Any text could be protected by copyright law and it is not always easy
to find suitable corpora that are free from copyright issues. Indeed,
the relationship between copyright of texts and their use in natural
language processing is complex
\cite{eckart-de-castilho-etal-2018-legal}.  Nonetheless, it pays off
to make some effort by searching for corpora that are free or in the
public domain \cite{oanc}. This makes it easier for other researchers
to work with it, in particular those that are employed by institutes
with lesser financial means.  The GMB only includes corpora from the
public domain \cite{gmb:lrec}.  Free parallel corpora are also
available via OPUS \cite{opus}. Other researchers take advantage of
vague legislation and distribute corpora quoting the right of fair use
\cite{postma:semeval}. Recently, crowd sourcing platforms such as
Figure Eight make datasets available, too (``Data For Everyone''),
under appropriate licensing.
\la{While targeting the public domain corpora, one might need to bear in mind the coverage of the corpora depending on the objectives of semantic annotation.}

\section{Freeze the corpus before you start}\label{sec:freeze}

Once you start your annotation efforts, it is a good idea to freeze
the corpora that will comprise your meaning bank.%
\footnote{Freezing the corpora already fixes certain data statements for your meaning bank, like curation rationale, language variety, and text characteristics. Communicating these data statements is important from an application point of view \cite{BenderFriedman:18}.}
In the GMB project \cite{gmb:lrec}, the developers were less strict in
maintaining this principle.  During the project they came across new
corpora, but after adding them to the GMB they were forced to fix and
validate annotations on many levels to get the newly added corpus up
to date and in sync with the rest. This problems manifests itself
especially for corpora that are constructed via a phenomenon-driven
annotation approach (Section~\ref{sec:phenomena}).

\section{Work with raw texts in your corpus}\label{sec:raw}

Keep the original texts as foundation for annotation. Never ever carry
out any semantic annotation on tokenised texts, but use stand-off
annotation on character offsets
(Section~\ref{sec:standoff}). Tokenisation can be done in many
different ways, and the \textit{atoms of meaning} certainly do not
correspond directly to words.  Most of the current conventions in
tokenisation are based on what has been used in (English)
syntax-oriented computational linguistics and can be misleading when
other languages are taken into consideration
(Section~\ref{sec:multi}). Moreover, if you use an off-the-shelf
tokeniser, you will find out soon that it makes mistakes---and
correcting those would break any annotations done at the word token
level. More likely, during your annotation project, you will find the
need to change the tokenisation guidelines to deal properly with
multi-word expressions (Section~\ref{sec:mwe}). 
In addition,
punctuation and spacing carry information that could be useful for
deep learning approaches, and their original appearance should therefore in one way or another
should be preserved.
An example: a ``New York-based'' company could be a new company based
in York, but the other interpretation is more likely. In an
NLP-processing pipeline, it is too late for syntax to fix this in a
compositional way---the tokenisation needs to be improved.

\section{Use stand-off annotation}\label{sec:standoff}

Stand-off annotation is a no-brainer as it offers a lot more
flexibility. 
It enables keeping annotations separate from the original raw text,
where ideally each annotation layer has its own file \cite{ide-romary-2006-representing,pustostubbs}. 
It is best executed with respect to the character
offsets of the raw texts in the corpus (Section~\ref{sec:raw}). A JSON or XML-based
annotation file can always be generated from this, should the demand
be there. Stand-off annotation is in particular advantageous in a
setting where several layers of annotation interact with each other
(typically in a pipeline architecture). This was extremely helpful in
the GMB \cite{gmb} where the document segmentation (sentence and word
boundaries) got improved several times during the project, without
having any negative effect on annotation occurring later in the
semantic processing pipeline (such as part-of-speech tagging and named
entity recognition).

\section{Consider manual annotation}\label{sec:manual}

Several meaning banks are created with the help of a grammar. The best
example here is the sophisticated English Resource Grammar
\cite{Flickinger:2000,Flickinger:2011} used to produce the treebanks,  Redwoods \cite{Oepen2004} and DeepBank \cite{Deepbank:2012}, annotated with English Resource Semantics (ERS) in a compositional way, by
letting the annotator pick the correct or most plausible analysis.
Similarly, the meaning representations in the GMB are system-produced
and partially hand-corrected \cite{gmb}, using a CCG parser
\cite{candc}. Likewise, the meaning representations in the PMB are
system-produced with the help of a CCG parser \cite{lewis:EMNLP2014}
and some of it is completely hand-corrected. In contrast,
the meaning representations of the AMR Bank are completely manually
manufactured---without the aid of a grammar---with the help of an
annotation interface and an extensive manual \cite{amr}.
\citet{bender-etal-2015-layers} argue that grammar-based meaning
banking requires less annotation guidelines, that it provides more
consistent analyses, and that it is more scalable. The downside of
grammar-based annotation is that several compound expressions are not
always compositional (negative and modal concord, postnominal genitives (``of
John's"), odd punctuation conventions, idioms), and that grammars with
high recall and precision are costly to produce (the impressive English Resource Grammar took about several years to develop, but it
is restricted to just one language).

\section{Make a friendly annotation interface}

Annotation can be fun (especially if gamification is applied, see
Section~\ref{sec:crowd}), but it can also be tedious. A good interface
helps the annotator to make high-quality annotations, to work
efficiently, and to be able to focus on particular linguistic phenomena. An annotation interface should be \la{web-based}
(i.e., any browser should support it), simple to use, and personalised.%
\footnote{\la{For more details about web-based collaborative annotation tools we refer to \citet{Biemannetal:2017}.}}
\la{The latter grants control over annotations of particular users}.
The ``Explorer''
\cite{explorer} introduced in the GMB and later further developed in
the PMB, has various search abilities (searches for phrases,
regular expressions, and annotation labels), a statistics page, a
newsfeed, and a user-friendly way to classify annotations as ``gold
standard''. The inclusion of a ``sanity checker'' helps to identify
annotation mistakes, in particular if there are several annotation
layers with dependencies. It is also a good idea to hook the
annotation interface up with a professional issue reporting system.

\section{Include an issue reporting system}

Annotators will sooner or later raise issues, have questions about the
annotation scheme, or find bugs in the processing pipeline. This is
valuable information for the annotation project and should not get
lost. The proper way to deal with this is to include a sophisticated
bug reporting system in the annotation interface.  For the GMB
\cite{gmb} and the PMB \cite{pmb}, 
the Mantis Bug Tracker\footnote{\url{https://www.mantisbt.org/}} 
was incorporated inside the Explorer \cite{explorer}. Besides Mantis there
are many other free and open source web-based bug tracking systems
available. A bug tracker enables one to categorize issues, assign them
to team members, \la{have dedicated discussion thread for each issue}, and keep track of all improvements made in a certain
time span (useful for the documentation in data releases).

\section{Be careful with the crowd}\label{sec:crowd}

Following the idea of Phrase Detectives \cite{phrasedetectives}, in
the GMB \cite{gmb} a game with a purpose \la{(GWAP)} was introduced to annotate
parts of speech, antecedents of pronouns, noun compound relations
\cite{BosNissim2015NoDaLiDa}, and word senses
\cite{VenhuizenBasileEvangBos2013IWCS}.  The quality of annotations
harvested from gamification was generally high, but the amount of
annotations relatively low---it would literally take years to annotate
the entire GMB corpus.  An additional problem with GWAPs is recruiting
new players: most players play the game only once, and attempts to make the game addictive could be irresponsible \cite{gamification}.
The alternative, engaging people by financially awarding them via
crowdsourcing platforms such as Mechanical Turk or Figure
Eight, solves the quantity problem \cite{pustostubbs}, but introduces
other issues including the question what a proper wage would be
\cite{fort-etal-2011-last} and dealing with tricksters and cheaters
\cite{buchholz}.

\section{Profit from lexicalised grammars}\label{sec:lex_grammar}

A lexicalised grammar gives an advantage in annotating syntactic
structure.
\la{In case of the compositional semantics, this also leads to automatic construction of the phrasal semantics.} 
This is because, in a lexicalised grammar, most of the
grammar work is done in the lexicon (there is only a dozen general
grammar rules), and annotation is just a matter of giving the right
information to a word (rather than selecting the correct
interpretation from a possibly large set of parse trees). In the PMB a lexicalised grammar is used: Combinatory Categorial
Grammar (CCG, \citealt{steedman:sp}), and the core annotation layers for each word token
are a CCG category, a semantic tag \cite{AbzianidzeBos2017IWCS}, a lemma, and a word sense.
Annotating thematic roles (Section~\ref{sec:roles}) is also convenient in a lexicalised grammar environment \cite{BosEvangNissim2012}.
\la{Finally, a lexicalised grammar coupled with compositional semantics facilitates annotation projection for meaning preserving translations and opens the door to multilingual meaning banking (Section\,\ref{sec:multi}).
Projection of meaning representation from one sentence to another is reduced to word alignment and word-level annotation transfer.
This type of projection is underlying the idea of moving from the monolingual GMB to the multilingual PMB.}

\section{Try to use language-neutral tools}\label{sec:neutral_tools}

Whenever possible, in machine-assisted annotation, get language
technology components that are not tailored to specific languages, because this increases portability of meaning processing components to other languages (Section\,\ref{sec:multi}).
The statistical tokeniser (for word and sentence segmentation) used in the
PMB is Elephant \cite{elephant}. The current efforts in multi-lingual
POS-tagging, semantic tagging \cite{AbzianidzeBos2017IWCS} and
dependency parsing are promising \cite{white-etal-2016-universal}.
In the PMB a categorial grammar is used to cover four languages (English, Dutch, German, and Italian), using the same parser and grammar, but with language-specific statistical models trained for the EasyCCG parser \cite{lewis:EMNLP2014}.
Related are grammatical frameworks designed for parallel grammar writing \cite{ranta:gf,Ben:Dre:Fok:Pou:Sal:10}.

\section{Apply normalisation to symbols}\label{sec:normalize_symbols}

Normalising the format of non-logical symbols (the predicates and individual constants, as opposed to logical symbols such as negation and conjunction) in meaning
representations decreases the need for awkward background knowledge
rules that would otherwise be needed to predict correct entailments.
Normalisation \cite{goot:phd} can be applied to date expressions
(e.g., the 24th of February 2010 vs. 24-02-2010 or dozens of variations on these), time expressions
(2pm, 14:00, two o'clock), and numerical expressions (twenty-four, 24,
vierundzwanzig; three thousand, 3,000, 3000, 3 000).  Compositional
attempts to any of the above mentioned classes of expressions are
highly ambitious and not recommended. Take, for instance, the Dutch
clock time expression ``twee voor half vier'', which denotes 03:28 (or
15:28)---how would you derive this compositionally in a \jb{computational}
straightforward way? Other normalisations for consideration are
expansion of abbreviations to their full forms, lowercasing proper
names, units of measurement, and scores of sports games. To promote
inter-operability between annotated corpora, it is a good idea to
check whether any standards are proposed for normalisation
\cite{pustostubbs}.

\section{Limit underspecification}

Underspecification is a technique with the aim to free the semantic
interpretation component from a disambiguation burden
\cite{reyle:udrs,bos:plu,mrs}.  In syntactic treebanks, however, the
driving force has been to assign the most plausible parse tree to a
sentence. This makes sense for the task of statistical (syntactic)
parsing.  The same applies to (statistical) semantic parsing: a corpus
with the most likely interpretation for sentences is
required. 
Moreover, it is not straightforward to draw correct
inferences with underspecified meaning representations \cite{reyle:reasoning}. So it makes sense, at least from the perspective of semantic annotation,
to produce the most plausible interpretation for a given sentence.
Consider the following examples. A ``sleeping bag'' could be a bag
that is asleep, but it is very unlikely (even in a Harry Potter
setting), so should be annotated as a bag designed to be slept in.  In
the sentence ``Tom kissed his mother'', the possessive pronoun could
refer to a third party, but by far the most likely interpretation is
that Tom's mother is kissed by Tom, and that reading should be
reflected in the annotation. Genuine scope ambiguities are relatively rare in ordinary text,
and it is questionable whether the representational overhead of underspecified scope is worth the effort given the low frequency of the phenomenon.
Nonetheless, resolving ambiguities is sometimes
hard, in particular for sentences in isolation. What is plausible for
one annotator is implausible for another. Finally, one needs to be careful, as annotation guidelines that
give preference for one particular reading (based on statistical
plausibility) have the danger of introducing or even amplifying bias.

\section{Beware of annotation bias}

Assigning the most likely interpretation to a sentence can also give
an unfair balance to stereotypes.  In the PMB, gender of
personal proper names are annotated. In many cases this is a
straightforward exercise.  But there are sometimes cases where the
gender of a person is not known. The disturbing distribution of male
versus female pronouns (or titles) strongly suggests that a female is
the least likely choice \cite{webster}. But following this statistical
suggestion only causes greater divide. The PMB annotation guidelines
for choosing word senses (Section~\ref{sec:senses}) are such that when it is
unclear what sense to pick, the higher sense (thus, the most frequent
one), must be selected. This is bad, because systems for word sense
disambiguation already show a tendency towards assigning the most
frequent sense \cite{postma:bias}. More efforts are needed to reduce
bias \cite{BCWS16}.

\section{Use existing resources for word senses}\label{sec:senses}

The predicate symbols that one finds in meaning representation are
usually based on word lemmas. But words have no interpretation, and a
link to concepts in an existing ontology \cite{cyc,babelnet} is
something that is needed to make the non-logical symbols in meaning
representations interpretable. In the AMR Bank, verbs are
disambiguated by OntoNotes senses \cite{amr}. In the PMB,
nouns, verbs, adjectives and adverbs are labelled with the senses of
(English) WordNet \cite{wordnet}.  Picking the right sense is
sometimes hard for annotators, sometimes because there is too little context, but also because the definitions of fine-grained senses are
sometimes hard to distinguish from each other
\cite{lopez-de-lacalle-agirre-2015-crowdsourced}.  Annotation
guidelines are needed for ambiguous cases where syntax doesn't help to
disambiguate: ``Swimming is great fun.'' (\texttt{swimming.n.01} or
perhaps \texttt{swim.v.01}?), ``Her words were emphasized.''
(\texttt{emphasized.a.01} or \texttt{emphasize.v.02}?).
WordNet's coverage
is impressive and substantial, but obviously not all words are listed (example:
names of products used as nouns) and sometimes it is inconsistent (for
instance, ``apple juice'' is in WordNet, but ``cherry juice'' is not).
Many WordNets exists for languages other than English
\cite{babelnet,multiwordnet}.

\section{Apply symbol grounding}

Symbol grounding \la{helps} to connect abstract representations of
meaning with objects in the real world or to unambiguous descriptions
of concepts or entities. This happens on the conceptual level with mapping words to
WordNet synsets or to a well-defined inventory of relations. Princeton
WordNet \cite{wordnet} lists several instances of famous persons but obviously the list is incomplete. The AMR Bank includes links from named entities to
wikipedia pages, but obviously not every named entity has a
wikipedia entry. To our knowledge, no other meaning banks apply
wikification.  Other interesting applications for symbol grounding are
GPS coordinates for toponyms \cite{Leidner:2008}, visualisation of
concepts or actions \cite{babelnet}, or creating timelines \cite{bamman-smith-2014-unsupervised}.

\section{Adopt neo-Davidsonian events}

It seems that in most (if not all) semantically annotated corpora a
neo-Davidsonian event semantics is adopted. This means that every
event introduces its own entity as a variable, and this variable can
be used to connect the event to its thematic roles. In the original
Davidsonian approach, an event variable was simply added to the
predicate introduced by the verb \cite{davidson:1967,kampreyle:drt} as a way to add modifiers 
(e.g., moving from \texttt{eat(x,y)} to \texttt{eat(e,x,y)} for a transitive use of \textit{to eat}).
In most modern meaning representations thematic roles are introduced
to reduce the number of arguments of verbal predicates to one, also known as the neo-Davidsonian tradition \cite{parsons:1990} (e.g., moving from \texttt{eat(e,x,y)} to \texttt{eat(e) AGENT(e,x) PATIENT(e,y)}).  
A direct consequence of a neo-Davidsonian design is the need for an inventory
of thematic roles.
But there is also an alternative, which is given a fixed arity to event predicates, of which some of them may be unused \cite{tacitus} when the context does not provide this information (e.g., for the intransitive usage of \textit{to eat}, still maintain \texttt{eat(e,x,y)} where \texttt{y} is left unspecified).

\section{Use existing role labelling inventories}\label{sec:roles}

A neo-Davidsonian approach presupposes a dictionary of thematic (or
semantic) role names. There are three popular sets available:
PropBank, VerbNet, and FrameNet.  PropBank \cite{propbank} proposes a
set of just six summarising roles: ARG0 (Agent), ARG1 (Patient), ARG2
(Instrument, Benefactive, Attribute), ARG3 (Starting Point), ARG4
(Ending Point), ARGM (Modifier). The interpretation of these roles are
in many cases specific to the event in which they participate.  The
AMR Bank adopts these PropBank roles \cite{amr}.
VerbNet has a set of about 25 thematic roles independently defined
from the verb classes \cite{verbnet}. A few examples are: Agent,
Patient, Theme, Instrument, Experiencer, Stimulus, Attribute, Value,
Location, Destination, Source, Result, and Material. The PMB adopts
the thematic roles of VerbNet.
FrameNet is organised quite differently. Its starting point is not
rooted in linguistics, but rather in real-world situations, classified
as frames \cite{framenet}. Frames have frame elements that can be
realised by linguistic expressions, and they correspond to the
PropBank and VerbNet roles.  There are more than a thousand different frames, and
each frame has its own specific role set (frame elements). For instance,
the Buy-Commerce frame has roles Buyer, Goods, Seller, Money, and so on.
There are also recent proposals for comprehensive inventories for
roles introduced by prepositional and possessive constructions
\cite{schneider}.
\la{In the PMB, we employ a unified inventory of thematic roles (an extension of the VerbNet roles) that is applicable to verbs, adjectives, prepositions, possessives or noun modifiers.}

\section{Treat agent nouns differently}

Agent and recipient nouns (nouns that denote persons performing or
receiving some action, such as employee, victim, teacher, mother,
cyclist, victim) are intrinsically relational \cite{booij:1986}. Modelling them like
ordinary nouns, i.e., as one-place predicates, can give rise to
contradictions for any individual that has been assigned more than one
role, because while you want to be able to state that a violin player
is not the same thing as a mother, a person could perfectly be a
mother and a violin player at the same time. Moreover, a fast cyclist
could be a slow driver.  Incorrect modeling can furthermore lead to
over-generation of some unmanifested relations (for instance, if Butch
is Vincent's boss and Mia's husband, a too simple model would predict
that Butch is also Vincent's husband and Mia's boss.  In the AMR Bank
\cite{amr} agent nouns are decomposed (e.g., an ``investor'' is a
person that invests). In the PMB agent nouns introduce
a mirror entity (e.g. an ``investor'' is a person with the
role of investor).

\section{Beware of geopolitical entities}

Names used to refer to geopolitical entities (GPEs) are a real pain in
the neck for semantic annotators.  How many times did we change the
annotation guidelines for these annoying names! The problem is that
expressions like ``New York'', ``Italy'', or ``Africa'' can refer to
locations, their governments, sport squads that represent them, or the
people that live there (and in some case to multiple aspects at the
same time, as in ``Italy produces better wine than France''). 
This instance of systematic polysemy manifests itself for all classes of 
GPE, including continents, countries, states, provinces, cities, and so on. Detailed instructions
for annotating GPEs can be found in the ACE annotation guidelines
\cite{ace}.

\section{Give scope to negation}\label{sec:negation}

Sentence meaning is about assigning truth conditions to propositions
(Section~\ref{sec:tarski}).  Negation plays a crucial role here---in
fact, the core of semantics is about negation, identifying whether a
statement is true of false.  Negation is a semantic phenomenon that
requires scope, in other words, it cannot be modelled by simply
applying it as a property of an entity.  It should be clear---explicit
or implicit---what the scope of any negation operator is, i.e. the
parts of the meaning representation that are negated.  The GMB, PMB
and DeepBank \cite{Deepbank:2012} assign proper scope to negation (the latter with the help
of underspecification). In AMR Bank negation is modelled with the help
of a relation, and this doesn't always get the required interpretation
\cite{Bos2016CL}.  Negation can be tricky: negation affixes
(Section~\ref{sec:tests}) require special care, negative concord
(Section~\ref{sec:manual}) and neg raising \cite{liu-etal-2018-negpar} are challenges for compositional approaches to meaning construction.

\section{Pay attention to compound words}\label{sec:mwe}

In the GMB \cite{gmb} we largely ignored multi-word expressions (MWEs),
believing that compositionality would eventually do away with
it. Except it doesn't.  MWEs come in various forms, and require
various treatments \cite{SagMWE:2002}. Think about proper names (names of persons,
companies, locations, events), titles and labels (of people, of books,
chapters, of songs), compounds, phrasal verbs, particle verbs, fixed phrases, and idioms.
Consider for instance ``North and South Dakota'', it is quite a
challenge to derive the representation state(x) \&
name(x,'North-Dakota') in a compositional way. And many compounds are
not compositional (``peanut butter'' is not butter, and ``athlete's
foot'' is not a body part but a nasty infection).  
It is hard to decide where to draw the line between a compositional and non-compositional approach to multi-word expressions. Even though ``red wine'' is written in English with two words, in German it is
written in one word (``rotwein''). 
WordNet \cite{wordnet} lists 
many multi-word expressions and could be used as a resource to decide
whether a compounds is analysed compositionally or not.
In the PMB, titles of songs or other artistic works are treated as a
single token (because they are proper names), which works fine for
``Jingle Bells'' but becomes a bit awkward and uncomfortable with longer titles such as Lennon and McCartney's  ``Lucy
in the Sky with Diamonds'', or Pink Floyds's ``Several Species of Small Furry Animals Gathered Together in a Cave and Grooving With A Pict''.
It is quite unfair  and unrealistic to expect
the tokeniser to recognise this as a multi-word expression. The
alternative, applying some reinterpretation after having first carried
out a compositional analysis, puts a heavier burden on the
syntax-semantics interface.
The bottom line is that MWEs form a wild bunch of expressions for which a general modelling strategy covering all types does not seem to exist. There also seems to be a connection with quotation \cite{Quotation:2014}.

\section{Use inference tests in design}\label{sec:tarski}
\label{sec:tests}

The driving force to motivate \la{how to shape or} what to include in
a meaning representation should be textual entailment or 
contradiction checks
(this is a practice borrowed from formal semantics).  For instance,
when designing a meaning representation for adjectives, the meaning
for ``ten-year-old boy'' should not imply that the boy in question is
old. Likewise, the meaning representation for ``unhappy'' should not
be the same as that for ``not happy'', because the
meanings of these expressions are not equivalent (as ``Bob is not
happy'' doesn't entail ``Bob's unhappy''---Bob can be both not happy
and not unhappy---even though the entailment holds in the reverse
direction: if Bob is unhappy, he is not happy).  Similarly, the meaning representation for ``Bologna is
the cultural capital of Italy'' should not lead to the incorrect
inference that ``Bologna is the capital of Italy''. 
In addition, or as alternative to inference checks,
is applying the method of model-theoretic interpretation \cite{blackburnbos:2005} 
when designing meaning representations.  
It should be clear what a representation actually means, in other words,
under which conditions it is true or false. A formal way of defining
this is via models of situation, and a satisfaction definition that
tells us, given a certain situation, whether a statement holds or
doesn't. This method was introduced by the logician Tarski \cite{tarski}. It bears similarities with posing a query to a relational database.
The method forces you to make a strict distinction between logical (negation, disjunction, equality) and non-logical symbols (the predicates and individual constants in your meaning representation).

\section{Divide and conquer}\label{sec:phenomena}

Do not try to do model all semantic phenomena the first time
around. There are just too many.  Some good candidates to put on hold are plurals, tense, aspect, focus, presupposition
(see Section~\ref{sec:presup}), and generics (more in  Section~\ref{sec:generics}), because a proper treatment of these phenomena requires a lot more than a basic predicate-argument structure. 
A strict formalisation of plurals quickly leads to complicated representations \cite{kampreyle:drt}, leading to compromising approximations in the AMR Bank \cite{amr} or PMB \cite{pmb}.
In the GMB \cite{gmb} and the AMR Bank tense is simply ignored. 
Annotating aspect is complex---for instance,
the use of the perfect differs enormously even between closely related languages such
as English, Dutch, and Italian \cite{van-der-klis-etal-2017-mapping}. 
These complications lead to a simple annotation model in the PMB where
tense is reduced to a manageable set of  three tenses: past, present and future. 
There are, therefore, a lot of
interesting problems left for the second round of semantic annotation!

\section{Put complex presuppositions on hold}\label{sec:presup}

Presuppositions are propositions that are taken for granted. Several
natural language expressions introduce presuppositions. These
expressions are called presuppositions triggers. (For instance, ``Mary
left, too.'' presupposes that someone else besides Mary left. Here
``too'' is the trigger of this presupposition.) There are many
different kinds of triggers, and many do not contribute to the meaning
of the sentence, but rather put constraints on the context.  The
question, then, is what to do with them in a meaning banking project.
Some classes of presupposition triggers, referring expressions
including proper names, possessive phrases, and definite descriptions,
can be treated in a similar way as pronouns, as is done in the GMB and the PMB, following \citet{bos:cl}. 
Yet there are
other classes of triggers that are notoriously hard to represent, \jb{because they require some "copying" of large pieces of meaning representation}, interact with focus, and require non-trivial semantic composition methods. To
these belong implicative verbs (manage), focusing adverbs (only,
just), and repetition particles (again, still, yet, another).  For
instance, although in the PMB a sentence like ``The crowd applauded again.'' is
the presupposition trigger, ``again'' is
semantically tagged as a repetition trigger, for now it doesn't perform any costly operations on the actual meaning representation. 
The first alternative, a meaning representation with two
different applauding events that are temporally related, is
complicated to construct. The second alternative, introducing
``again'' as a predicate, doesn't make sense semantically (what is the
meaning of ``again''?), or as an operator (again, how will it be
defined logically?) isn't attractive either. There are, currently no
good ways to deal with complex presupposition triggers, and more
research is needed here turning \jb{formal ideas \cite{KampRossdeutscher:2009} into practical solutions}.

\section{Respect elliptical expressions}

They are invisible, but omnipresent: elliptical expressions.
Comparative ellipsis is present in many languages (``My hair is
longer than Tom's'').  In English, verb phrase ellipsis occurs (``Tom
eats asparagus, but his brother doesn't.''), which is well studied
\cite{dalrymple:ellipsis}, and annotated corpora exist as well
\cite{BosSpenader2011}.  Dutch and German exhibit a large variety of
gapping cases (``Tom isst Spargel, aber sein Bruder nicht.'').
Italian is a language with pro-drop (``Ho fame'', i.e., (I) am hungry). 
Ellipsis requires a dedicated component in a pipeline architecture.
In the PMB the inclusion of an ellipsis
layer has been postponed for the benefit of other components, features, and efforts. As a
consequence, a growing number of documents cannot be added to the gold
set because there isn't an adequate way of dealing with a missing
pronoun, an odd comparison expression, or an elided verb phrase. 

\section{Think about generics}\label{sec:generics}

Generic statements and habituals are hard to model straightforwardly
in first-order logic \cite{carlson:phd}. 
The sentence ``a lion is strong'' or ``a dog has
four legs'' is not about a particular lion or dog, nor is it about all
dogs or lions. The inventor of ``the typewriter'' was not the inventor
of a particular typewriter, but of the typewriter concept in
general. Such generic concepts are also known as \textit{kinds} in the literature \cite{reiter-frank-2010-identifying}. 
It is not impossible to approximate this in first-order
logic, but it requires an ontological distinction between entities
denoting individuals and entities denoting concepts (kinds). A further
question is how tense should be annotated in habitual sentences, as in
``Jane used to swim every day'' (in some period in the past, Jane swam
every day) or ``Jane swims every day'' (in the current period, Jane
swims every day).
To our knowledge, none of the existing meaning banks have a satisfactory
treatment of generics, 
even though techniques have been proposed to detect generics \cite{reiter-frank-2010-identifying,friedrich-pinkal-2015-discourse}.
Recent proposals try to change this situation \cite{donatelli-etal-2018-annotation}.

\section{Don't try to be clever}

The English verb ``to be'' (and its counterpart in other Germanic languages) is a semantic nuisance. When used as an auxiliary
verb---including predicative usages of adjectives---there isn't much
to worry about it, as it only semantically contributes tense
information. However, when used as a copula it can express identity,
locative information, or predications involving nouns.  From a logical
perspective, it might seem attractive to use equality in these cases
and interpret ``to be'' logically rather than lexically,
\cite{blackburnbos:2005}, but this makes it impossible to include
tense information, unless equality is \jb{(non-standardly)} viewed as a
three-place relation. There are various senses for ``be'' in WordNet,
and it makes pragmatically sense to use these: ``This is a good idea''
(sense 1), ``John is the teacher'' (sense 2), ``the book is on the
table'' (sense 3), and so on. A similar story can be told for ``to
have'' in expressions like ``Mary has a son'', where the first attempt
in the PMB was to analyse ``to have'' in such possessive constructions
as logical, i.e. only introducing tense information, and coerce the
relational noun ``son'' into a possessive interpretation. This was
soon abandoned due to complications in composition.

\section{Don't focus on just one language}\label{sec:multi}

Most meaning banks consider just one language, and usually this is
English.  This is understandable, as English is the current scientific
language, but it is also risky, because when designing meaning
representation decisions could be made that work for English but not for
other languages.  Phenomena such as definite descriptions, ellipsis,
possessives, aspect, and gender, behave even in closely related
languages quite differently from each other.
Dealing with multiple languages is, without any doubt, harder, but if
one takes several languages into account at the same time the result
is more likely to be more language-neutral meaning
representations. And that's what meanings should be, they are abstract
objects, independent of the language used to expressed them. Of
course, there are concepts that can be expressed in certain languages
with a single word that other languages are not capable of, but the
core of meaning representations should be agnostic to the source
language. A good starting point is to work with typologically-related
languages. 
An efficient annotation technique to cover multiple languages is
\textit{annotation projection}
\cite{Evang2016COLING,liu-etal-2018-negpar}. This requires a parallel
corpus and automatic word alignment, and existing semantic annotations for at least one language.

\section{Measure meaning discrepancies}\label{sec:measure}

A large part of the users of semantically annotated corpora are from 
the semantic parsing area, and they need to be able to measure and
quantify their output with respect to gold standard meanings.  The
currently accepted methods are based on precision and recall on the
components of the meaning representation by converting them to triples or clauses
\cite{allen:step2008,dridan-oepen-2011-parser,cai-knight-2013-smatch,Noord2018LREC,ulf}.  
In a parallel corpus setting, such evaluation measures can also be used to
compare the meaning representation of a source text and its
translation \cite{saphra-lopez-2015-amrica}. 
This is done in the PMB, where a non-perfect
meaning match between source and target helps the annotator to
identify possible culprits. It is important to note that most of
these matching techniques check for syntactic equivalence, and don't
take semantic equivalence into account---the same meaning could be
expressed by syntactically different representations. The approach by
\citet{Noord2018LREC} applies normalisation steps for word senses to
make matching more semantic. 

\section*{Acknowledgments}

We would like to thank the two anonymous reviewers for their comments---they helped to improve this paper considerably.  Reviewer 1 gave us valuable pointers to the literature that we missed, and spotted many unclear and ambiguous formulations. Reviewer 2 was disappointed by the first version of this paper--- we hope s/he likes this improved version better.
This work was funded by the NWO-VICI grant “Lost in Translation  Found in Meaning” (288-89-003).

\bibliography{comsem}
\bibliographystyle{acl_natbib}

\end{document}